\documentclass[10pt,conference]{IEEEtran} 
\usepackage{cite}
\usepackage{amsmath,amssymb,amsfonts}

\usepackage{graphicx}
\usepackage{textcomp}
\usepackage{xcolor}
\usepackage{booktabs}
\usepackage{multirow}
\usepackage[table]{xcolor}
\usepackage{hyperref}
\usepackage{booktabs}
\usepackage{tabularx}
\usepackage{algorithm}
\usepackage{algpseudocode}
\usepackage{multirow} 
\usepackage{makecell}
\definecolor{rowgray}{gray}{0.92} 

\definecolor{myyellow}{RGB}{255,217,102}
\definecolor{myblue}{RGB}{143,170,220}
\definecolor{mygreen}{RGB}{169,209,142}
\makeatletter
\newcommand{\linebreakand}{%
  \end{@IEEEauthorhalign}
  \hfill\mbox{}\par
  \mbox{}\hfill
  \begin{@IEEEauthorhalign}
}
\makeatother
\IEEEoverridecommandlockouts

\def\BibTeX{{\rm B\kern-.05em{\sc i\kern-.025em b}\kern-.08em
    T\kern-.1667em\lower.7ex\hbox{E}\kern-.125emX}}
\begin{document}

\title{When LLMs Over-Answer: Measuring and Mitigating Quality Issues in LLM-Based Hardware Description Language Question Answering
}

\author{
\IEEEauthorblockN{Ziteng Hu}
\IEEEauthorblockA{\textit{Micro-Electronics Research Institute,}\\
\textit{Hangzhou Dianzi University}\\
Hangzhou, Zhejiang, China
\\ zitenghu@hdu.edu.cn}
\and
\IEEEauthorblockN{Jiachi Chen}
\IEEEauthorblockA{\textit{The State Key Laboratory of Blockchain }\\
\textit{and Data Security, Zhejiang University}\\
Hangzhou, Zhejiang, China
\\ chenjiachi@zju.edu.cn}
\and
\IEEEauthorblockN{Wenhao Lv}
\IEEEauthorblockA{\textit{Micro-Electronics Research Institute, }\\
\textit{Hangzhou Dianzi University}\\
Hangzhou, Zhejiang, China
\\ 241050106@hdu.edu.cn}

\linebreakand

\IEEEauthorblockN{Huan Zhang}
\IEEEauthorblockA{\textit{College of Information Science }\\
\textit{and Engineering, Hunan Normal University}\\
Changsha, Hunan, China
\\ zhanghuancs0123@gmail.com}
\and
\IEEEauthorblockN{Yingjie Xia\IEEEauthorrefmark{1}%
\thanks{\IEEEauthorrefmark{1} Corresponding author.}
}
\IEEEauthorblockA{\textit{Micro-Electronics Research Institute, }\\
\textit{Hangzhou Dianzi University}\\
Hangzhou, Zhejiang, China
\\ xiayingjie@zju.edu.cn}
}

\maketitle


\begin{abstract}
The rapid advancement of large language models (LLMs) has led practitioners to increasingly rely on them for answering questions about hardware description languages (HDLs). 
Because HDL is ultimately synthesized into physical hardware, an imprecise or redundant answer can propagate into timing violations or non-synthesizable logic that surface only late in the design flow, making the quality of HDL answers especially consequential.
However, the quality of LLM-generated responses, particularly in comparison with answers provided by human experts, remains unclear. To investigate this question, we collect 6,246 HDL Q\&A posts with accepted answers from Stack Overflow and curate them into a dataset, organized into a taxonomy of four main categories (Conceptual, Debugging, Generation, and Optimization) and ten subcategories. 
Using this dataset, we design a user study conducted with 19 HDL engineers with 1--3 years of experience. Our findings reveal a pervasive \emph{over-answering} tendency: LLMs supply correct content but bury it under redundant alternatives (65.7\%) and verbose padding (69.1\%), while nearly half of answers (49.0\%) fail to fully align with expert answers—yet participants still preferred LLM responses for readability (58.3\%).
Motivated by these findings, we propose a multi-agent framework for improving LLM-based HDL question answering. This framework employs multi-role agent debate to eliminate redundancy from core content and category-specific refinement to improve the conciseness of non-core content. We evaluate answer quality using an LLM-as-Judge and two structural metrics: the number of core answers, which reflects redundancy since LLMs often provide multiple alternative solutions, and the length of non-core content, which reflects verbosity.
Evaluated on the four mainstream LLMs, our framework increases the average core-answer quality score from 3.71 to 4.67 (+0.96) and the non-core content quality from 3.72 to 4.23 (+0.51), on a five-point scale. Meanwhile, it reduces the number of core answers by 37\% and non-core content length by 31\% on average. These results show that task-aware refinement can produce substantially more concise and focused HDL answers while improving their core technical quality.

\end{abstract}
\section{Introduction}

\begin{figure}[!t]
    \centering
    \includegraphics[width=1\columnwidth]{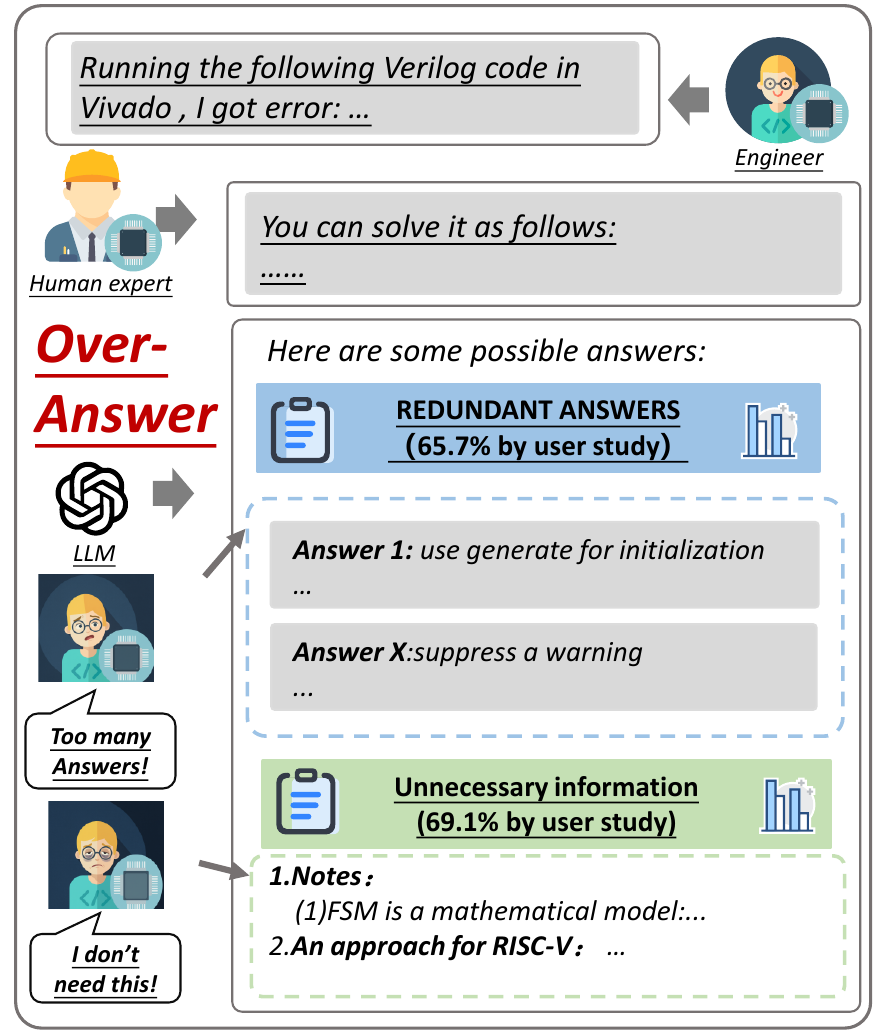} 
    \caption{An example of the quality gap between LLM-generated answers and accepted answers provided to a Verilog question. The human expert converges on a single decisive solution, whereas the LLM hedges with multiple alternative answers (redundancy, 65.7\% by our user study) and pads the response with unnecessary background and summaries (verbosity, 69.1\%).}
    \label{fig:motivation}
    \vspace{-1em}
\end{figure}

Hardware description languages (HDLs) play a central role in hardware design, as the code written in these languages directly determines the correctness and behavior of register-transfer-level (RTL) systems~\cite{1364_1}.
Engineers have traditionally relied on Question\&Answer (Q\&A) platforms such as Stack Overflow~\cite{stack} to address design, debugging, and optimization challenges. However, with the emergence of large language models (LLMs), an increasing number of engineers are turning to LLMs for assistance\cite{assessing,pansurvey}. As LLMs become increasingly capable of generating fluent answers, an important question arises: \textit{How do LLM-generated answers differ from accepted answers provided by human experts in real-world HDL question answering?} Fig.~\ref{fig:motivation} gives an example of a difference in a Verilog debugging question. The human expert provides a single, targeted solution, whereas the LLM returns several answers followed by background information that is unnecessary for resolving the problem. We refer to this combination of redundant alternatives and verbose padding as \emph{over-answering}---the LLM does not fail to answer, it answers too much.

However, evaluating the quality of the answer of a question is not easy. Existing work on LLMs for HDLs primarily targets code generation~\cite{verilogcoder,rtlcoder} and functional correctness, typically evaluated through simulation~\cite{verilogeval} or synthesis-based metrics~\cite{rtllm,betterv}. However, real-world HDL Q\&A is inherently open-ended: even a functionally correct answer may be of limited value to an engineer if it is redundant or inconsistent with expert practices. Answer quality therefore depends not only on functional correctness but also on several aspects, such as redundancy, conciseness, readability, and engineering usability~\cite{iso}---dimensions that remain largely underexplored in the HDL setting.

To address this gap, this paper is organized into two parts. 

\textbf{User Study.} We conduct a user study to characterize the quality gap between LLM-generated and accepted answers provided by human experts on Stack Overflow. We construct a dataset comprising 6,246 Stack Overflow HDL questions and their corresponding accepted answers. The questions are organized into four task categories (Conceptual, Debugging, Generation, and Optimization) and ten fine-grained subcategories as prior work on LLMs for HDL~\cite{llm4eda}. Using this dataset, 19 Verilog and hardware engineers with 1--3 years of experience evaluate LLM-generated and accepted answers along four dimensions: readability, redundancy, conciseness, and consistency with the accepted answer.

The user study reveals a multi-dimensional quality gap. We find that 65.7\% of LLM responses exhibit redundancy in their \emph{core answer}---the part that directly answers the question---by providing multiple alternative solutions, and that 69.1\% are padded with \emph{non-core content}---background, restatement, and summaries---that the question did not call for. Moreover, in 49.0\% of cases, the LLM answer does not fully cover the accepted answer. Despite these limitations, participants preferred the LLM-generated answer for readability in 58.3\% of the evaluated questions, compared with the corresponding accepted answer.
These findings suggest that LLMs are not simply worse than human experts; rather, they exhibit a different quality profile: more readable and explanatory, but also more redundant, more verbose, and less aligned with the accepted answer.

\textbf{Task-aware Multi-Agent Framework.} Motivated by the above findings, we propose a task-aware multi-agent refinement framework. Guided by the task category, the framework first profiles each question to infer its task type, user intent, and verification needs, and then uses multi-role agent debate to eliminate redundancy from the core answer and category-specific refinement to reduce verbosity in the non-core content. Because the core answers are examined jointly, the debate prunes not only incorrect candidates but also derivative ones that merely repackage a stronger solution, while keeping task-critical material and removing only dispensable background and summaries.

We evaluate our framework on four mainstream LLMs. The framework raises the average core-answer quality from 3.71 to 4.67 ($+0.96$) and the non-core content quality from 3.72 to 4.23 ($+0.51$) on a five-point scale, while reducing the number of core answers by 37\% and the non-core length by 31\%. These gains hold across all four backbone models and four task categories, indicating that the improvement stems from the structured decomposition and debate rather than from any single model or task. A further consistency analysis shows that refinement preserves or slightly improves consistency with the accepted answer, confirming that the reductions come from discarding non-essential content rather than cutting substantive information.

In summary, this paper makes the following contributions:
\begin{itemize}
    \item We conduct an empirical user study with 19 Verilog and hardware engineers, revealing a multi-dimensional quality gap.
    \item We propose a task-aware multi-agent refinement framework that eliminates redundancy in the core answer and reduces verbosity in the non-core content, improving answer quality while preserving consistency with the accepted answer.
    \item We release our dataset consists of 6,246 real-world HDL Q\&A pairs with accepted answers and the source code of the framework at 
    https://github.com/ZitengHu/HDLQA
    to facilitate further research.
\end{itemize}

\section{Background}
\label{background}

\subsection{Hardware Description Language Development}
Hardware description languages (HDLs) model, simulate, and synthesize digital circuits at the register-transfer level (RTL)~\cite{verilog}. The two most widely adopted, Verilog and VHDL, are standardized as IEEE~1364~\cite{1364} and IEEE~1076~\cite{IEEE1076}, with SystemVerilog extending Verilog as a design-and-verification superset. Unlike general-purpose code, HDL is ultimately synthesized into physical hardware, imposing strict requirements on correctness, timing, and synthesizability~\cite{1364_1}. A typical workflow spans design, simulation, debugging, and optimization~\cite{verilog}, tasks that differ markedly in their cognitive demands---from language semantics for conceptual questions, to causal reasoning for debugging~\cite{rtlfixer,veridebug}, synthesizable standards-compliant code for generation~\cite{verilogeval,rtllm}, and timing--resource trade-off analysis for optimization~\cite{codev}.



\subsection{HDL Question Answering in Practice}
Because HDL is synthesized into hardware, an incorrect answer can propagate into timing violations or non-synthesizable logic that surface only late in the design flow~\cite{verilog, nonblocking}. Engineers have long relied on Q\&A platforms such as Stack Overflow, where an \emph{accepted answer} endorsed by the questioner serves as a peer-validated reference~\cite{stack}. HDL questions are tightly coupled to tool behavior, standards compliance, and hardware semantics~\cite{1364, llm4eda}, so a useful answer must not only be correct but also align with established practice. As engineers increasingly turn to LLMs in place of these platforms~\cite{pansurvey, assessing}, an answer's value depends not only on functional correctness but on whether it is concise, focused, and consistent with how experts respond---qualities that code-generation metrics like VerilogEval and RTLLM do not capture~\cite{verilogeval, rtllm}. This motivates studying the \emph{communicative} quality of HDL answers as a distinct problem.

\begin{table*}[t]
\centering
\caption{Distribution and Definition of HDL Q\&A Tasks}
\label{tab:category_distribution}
\renewcommand{\arraystretch}{1.2}
\setlength{\tabcolsep}{5pt}
\begin{tabularx}{\textwidth}{ll X cc}
\toprule
\textbf{Main Category} & \textbf{Subcategory} & \textbf{Description} & \textbf{\# Instances} & \textbf{Rank} \\
\midrule
\rowcolor{gray!20}
Conceptual (2,009) & syntax-explanation & Language constructs, semantics, or coding rules~\cite{nonblocking}. & 1,450 & 2 \\
\rowcolor{gray!20}
 & tool-usage & Operating EDA toolchains and interpreting their messages~\cite{verilog}. & 559 & 4 \\
\midrule
Debugging (3,139) & functional-error & Compiles, but behavior deviates from the specification~\cite{veridebug}. & 1,452 & 1 \\
 & syntax-error & Violates HDL grammar; fails to compile or elaborate~\cite{rtlfixer}. & 1,189 & 3 \\
 & synthesis-error & Simulates correctly but is not synthesizable~\cite{nonblocking}. & 498 & 5 \\
\midrule
\rowcolor{gray!20}
Generation (630) & logic-generation & Control-oriented or structural RTL (e.g., FSMs, decoders)~\cite{verilogeval}. & 488 & 6 \\
\rowcolor{gray!20}
 & arithmetic-generation & Datapath or arithmetic hardware (e.g., adders, multipliers)~\cite{rtllm}. & 142 & 9 \\
\midrule
Optimization (468) & readability-optimization & Refactoring for clarity without changing function~\cite{verilog}. & 191 & 7 \\
 & ppa-optimization & Improving power, performance, or area~\cite{betterv}. & 152 & 8 \\
 & security-optimization & Hardening against hardware-level threats~\cite{secv,secfsm}. & 125 & 10 \\
\bottomrule
\end{tabularx}
\end{table*}

\subsection{HDL Task Type}
Prior work on LLMs for HDL has organized hardware tasks along recurring lines~\cite{llm4eda}---specification-to-RTL generation~\cite{verilogeval,rtllm}, debugging and repair~\cite{rtlfixer,veridebug}, broader workflows spanning verification and question answering~\cite{cvdp}, and optimization for power, performance, and area~\cite{betterv,rtlcoder}. Building on these distinctions, we group HDL questions into four main categories---\emph{conceptual}, \emph{debugging}, \emph{generation}, and \emph{optimization}---spanning ten subcategories, whose definitions and distribution are summarized in Table~\ref{tab:category_distribution}.


\section{User Study: Quality Gaps Between LLMs and Human Experts}

\label{sec:user_study}

\begin{figure}[h]
    \centering
    \includegraphics[width=1\columnwidth]{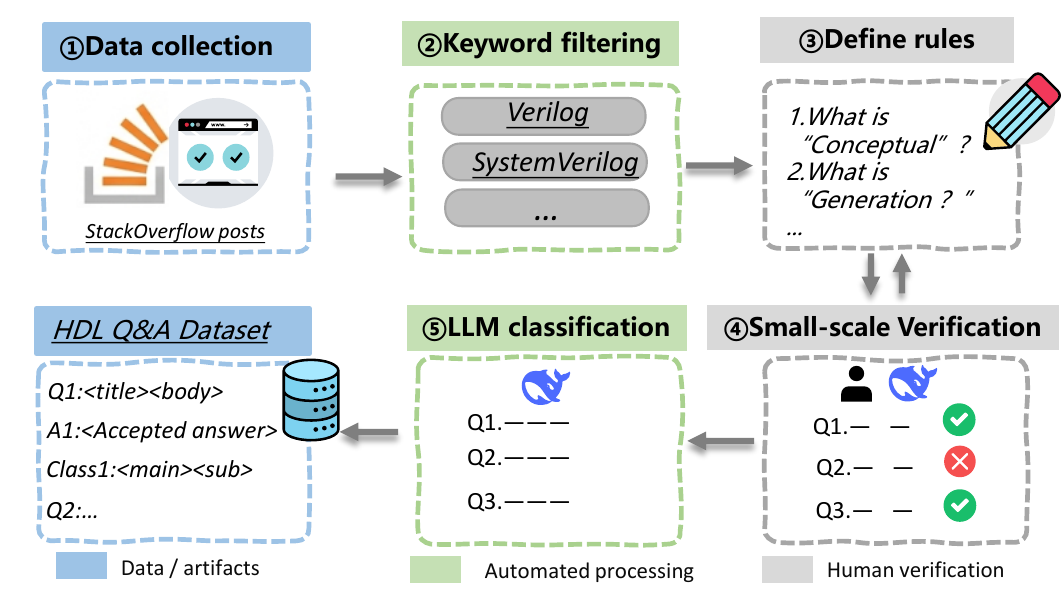} 
    \caption{Pipeline for constructing and annotating the HDL Q\&A dataset. We first collect HDL-related posts. Then, we design an initial taxonomy and rules, validate and refine them via human verification, and then apply the finalized rules for large-scale LLM-based classification.}
    \label{fig:dataset}
    \vspace{-1em}
\end{figure}

To study the quality differences between LLM-generated and accepted answers in HDL Q\&A posts, we follow the pipeline in Fig.~\ref{fig:dataset}. Specifically, we collect Stack Overflow posts with accepted answers, filter HDL-related ones by keyword (e.g., Verilog, SystemVerilog, VHDL), and design a task taxonomy over Conceptual, Debugging, Generation, and Optimization categories with annotation rules for LLM-based classification. We validate the taxonomy and rules on a stratified sample of 363 questions, refine ambiguous rules from the observed disagreements, and apply the refined classifier to all 6{,}246 questions for downstream analysis.

\subsection{Data Collection and Processing}

\subsubsection{Stack Overflow Posts}
We use the public Stack Overflow data dump, a cumulative snapshot containing all posts from the site's inception up to the dump release date (the \texttt{[2025-06-30]} release). From this snapshot we retain only questions that have an accepted answer provided by human experts, since accepted answers provide a practical reference for how developers resolve real problems. For each such post, we extracted the question title, body, tags, code snippets, accepted answer, answer/question scores, creation time, and other metadata.

\subsubsection{Keyword-based Filtering}
We filtered HDL-related posts using a keyword list of HDL-specific terms (e.g., \textit{verilog}, \textit{VHDL}, \textit{HDL}) and hardware-design terms (e.g., \textit{FSM}, \textit{latch}, \textit{RTL}), iteratively removing keywords that appeared mainly in irrelevant contexts to reduce false positives. The resulting dataset spans both dominant HDL families: Verilog/SystemVerilog and VHDL make up 53.4\% and 43.2\% of all records, the remaining 3.4\% being cross-family or untagged posts (full distribution in the supplementary material), confirming it is representative of mainstream HDL practice rather than Verilog-specific. The filtering and data-processing scripts were implemented with the assistance of a generative AI tool.

\subsubsection{Define rules for LLM}
To understand the types of HDL questions discussed by developers, we first defined a classification rules for LLM-based labeling. The rules were designed according to common HDL development tasks and hardware-design concepts observed in Stack Overflow posts. We used DeepSeek-V4 as the classifier throughout the annotation pipeline, following recent evidence that LLMs can perform text-annotation and classification tasks with reliability comparable to human annotators~\cite{chatgpt2,chatgpt1}.

\subsubsection{Small-scale Verification}
Before full-scale annotation, we conducted a small-scale human verification to assess the reliability of the classification rules. As manually checking all 6{,}246 questions is impractical, we determined the sample size using Cochran's formula~\cite{Sampling} with a 95\% confidence level ($Z = 1.96$), $p = 0.5$, and a 0.05 margin of error, yielding $n_0 \approx 385$; the finite-population correction for $N = 6{,}246$ gives $n \approx 363$. We drew these samples via stratified random sampling~\cite{Sampling} across the four task strata, proportional to each stratum's size ($n_i = 363 \times N_i / 6246$).

Treating the 363 human-verified samples as ground truth, the classifier achieved \textbf{93.94\%} category-level and \textbf{92.29\%} subcategory-level accuracy. Because the sample size was fixed by Cochran's formula, these estimates generalize to the full dataset: the Wilson score intervals~\cite{wilson1927probable} give 95\% CIs of $[91.0\%, 96.0\%]$ and $[89.1\%, 94.6\%]$ for category and subcategory accuracy. This matches prior evidence that LLMs rival human annotators on text classification~\cite{chatgpt1,chatgpt2}, and errors are dominated by near-miss subcategory confusions rather than cross-category misclassification, so the category-level analyses underlying our findings rest on reliable labels.

\subsubsection{LLM Classification}
After finalizing the classification rules, we applied the LLM classifier to all 6,246 HDL-related questions in our dataset. For each question, the LLM produced a structured classification result, including the main category, subcategory, and reason. We then parsed the model outputs and calculated the distribution of different HDL question task types as shown Table~\ref{tab:category_distribution}.
The classification results provide an overview of the real-world HDL questions that developers commonly ask about on Stack Overflow. They also serve as the basis for our later evaluation of LLM-generated answers.

\subsection{Study Design}

We designed a user study to compare LLM-generated answers with accepted answers provided by human experts. For each HDL question, participants were shown the original question and two candidate answers, denoted as Answer A and Answer B. One answer was generated by an LLM, and the other was the accepted answer provided by human experts. Participants evaluated the answers along multiple quality dimensions.

\subsubsection{LLM Answer Generation}
To obtain the LLM-generated answers used in the study, we prompted four LLMs spanning different model families and providers: Qwen3.5, DeepSeek-V4, GPT-5.4, and Gemini-3.1. Each question was assigned to one randomly selected model, which was queried under a zero-shot setting: the original Stack Overflow question (title and body, including any embedded code) was provided as input without additional examples, retrieval, or task-specific instructions, so as to reflect how practitioners typically query an LLM in practice. 

\subsubsection{Data Sample}
We use the same 363-question stratified sample. For each question, one of the four LLMs was randomly assigned to generate the answer, so the models are evaluated on disjoint subsets under a balanced design. We built 19 questionnaire versions, each completed by one participant and balanced across the four categories.

Each question was evaluated by a single annotator. Rather than collecting redundant labels on a small set of questions, this design prioritizes broader coverage across the 363 sampled questions and the four task categories. Because redundancy and conciseness are defined by relatively objective criteria, namely the presence of multiple alternative solutions and the presence of non-core content, single annotation provides a reasonable estimate of their prevalence. To assess reliability, a second annotator independently re-labeled a subset covering all four dimensions; inter-annotator agreement is substantial-to-almost-perfect (macro-average $91.5\%$, $\kappa=0.84$)~\cite{landis1977}, supporting the single-annotation design. Per-dimension figures are reported in the supplementary material.

\subsubsection{Quality Dimensions}
The questionnaire evaluates answer quality along four dimensions. For every item, each answer's identity (accepted vs. LLM) was shown beneath the title, and A/B order was randomized to remove position bias.

\begin{itemize}

\item \textit{\textbf{Consistency}} measures whether the LLM answer covers the core answer of the accepted answer, with four options: fully covered, partially covered, completely missing, or unable to judge. 

\item \textit{\textbf{Redundancy}} asks whether each answer contains repeated, overlapping or off-topic answer within its core part, with three options: yes, no, or unable to judge.

\item \textit{\textbf{Conciseness}} asks whether each answer avoids unnecessary background, summaries, or irrelevant extensions, with three options: yes, no, or unable to judge.

\item \textit{\textbf{Readability}} asks participants to choose which answer provides a better reading experience, with four options: A is better, B is better, tie, or unable to judge.

\end{itemize}

For redundancy and conciseness, participants judged Answer A and Answer B separately, choosing yes, no, or unable to judge for each. We did not include any question marked as unable to judge in the final statistics.

\subsection{Results}


\begin{table}[t]
\centering
\caption{User-study results across four evaluation dimensions (363 questions).}
\label{tab:userstudy}
\renewcommand{\arraystretch}{1.2}
\setlength{\tabcolsep}{6pt}
\small
\begin{tabular}{llcc}
\toprule
\textbf{Dimension} & \textbf{Outcome} & \textbf{Human expert} & \textbf{LLM} \\
\midrule
\rowcolor{gray!20}
Redundancy $\downarrow$  & flagged as redundant & 4.4\% & 65.7\% \\
Conciseness $\downarrow$ & flagged as verbose   & 5.8\% & 69.1\% \\
\midrule
\rowcolor{gray!20}
  & prefer Human expert & \multicolumn{2}{c}{35.1\%} \\
\rowcolor{gray!20}
  & prefer LLM          & \multicolumn{2}{c}{58.3\%} \\
\rowcolor{gray!20}
\multirow{-3}{*}{Readability} & tie & \multicolumn{2}{c}{6.6\%} \\
\midrule
\multirow{4}{*}{Consistency}
  & full coverage    & \multicolumn{2}{c}{50.7\%} \\
  & partial coverage & \multicolumn{2}{c}{41.3\%} \\
  & missing          & \multicolumn{2}{c}{7.7\%}  \\
  & unable to judge        & \multicolumn{2}{c}{0.3\%}  \\
\bottomrule
\end{tabular}
\end{table}

We report results across the four dimensions (Table~\ref{tab:userstudy}), then analyze task-category variation. Overall there is a clear multi-dimensional gap, most pronounced in redundancy, conciseness, and alignment with accepted answers.

\paragraph{Redundancy}

Redundancy is the most prominent issue: 65.7\% of LLM responses are flagged redundant versus 4.4\% for accepted answers (Table~\ref{tab:userstudy}), about 14.9$\times$ higher. It typically manifests as multiple alternative, overlapping, or off-topic answers that add no core information.

\begin{figure}[h]
    \centering
    \includegraphics[width=1\columnwidth]{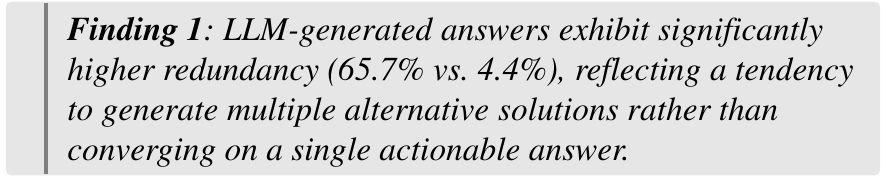} 
    \label{fig:redundancy}
    \vspace{-2em}
\end{figure}

\paragraph{Conciseness}

Conciseness degrades similarly---69.1\% of LLM responses are flagged verbose versus 5.8\% (Table~\ref{tab:userstudy}), a $\sim$12$\times$ increase---reflecting unnecessary background, extended summaries, and non-essential context beyond the core solution.

\begin{figure}[h]
    \centering
    \includegraphics[width=1\columnwidth]{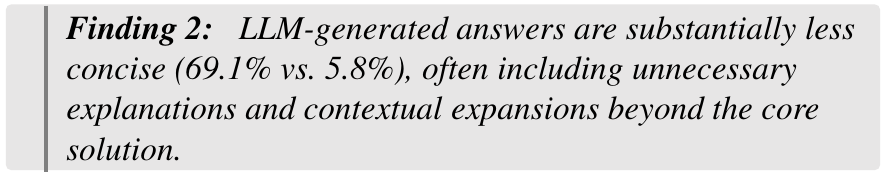} 
    \label{fig:conciseness}
    \vspace{-2em}
\end{figure}

\paragraph{Consistency}

49.0\% of LLM answers do not fully cover the accepted answer (41.3\% partial, 7.7\% missing; Table~\ref{tab:userstudy}). We call this the consistency-with-accepted-answer gap, but it should not be read as a direct correctness failure, since accepted answers may not be uniquely optimal.

\begin{figure}[h]
    \centering
    \includegraphics[width=1\columnwidth]{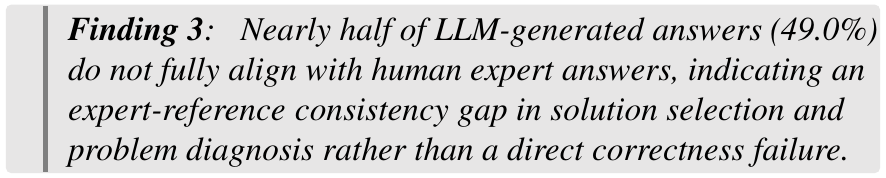} 
    \label{fig:consistency}
    \vspace{-2em}
\end{figure}

\paragraph{Readability}

LLM answers were preferred for readability in 58.3\% of questions versus 35.1\% for the accepted answer (Table~\ref{tab:userstudy}), a +23.2\% gap. LLMs thus read as more fluent and organized, though this advantage is largely surface-level rather than deeper engineering quality.

\begin{figure}[h!]

    \centering
    \includegraphics[width=1\columnwidth]{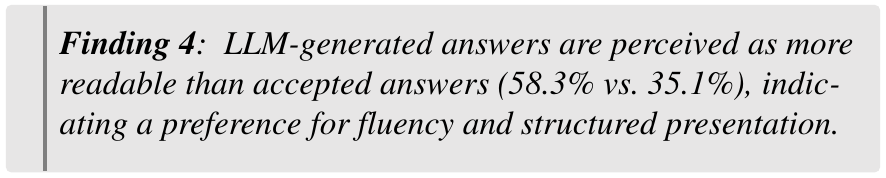} 
    \label{fig:readability}
    \vspace{-2em}
\end{figure}

Together, the redundancy and conciseness gaps describe a single underlying tendency---\emph{over-answering}: the model's core answer is present but accompanied by alternative solutions (redundancy) and non-essential padding (verbosity).

\subsubsection{Task-dependent effects}

\begin{figure}[h]
    \centering
    \includegraphics[width=1\columnwidth]{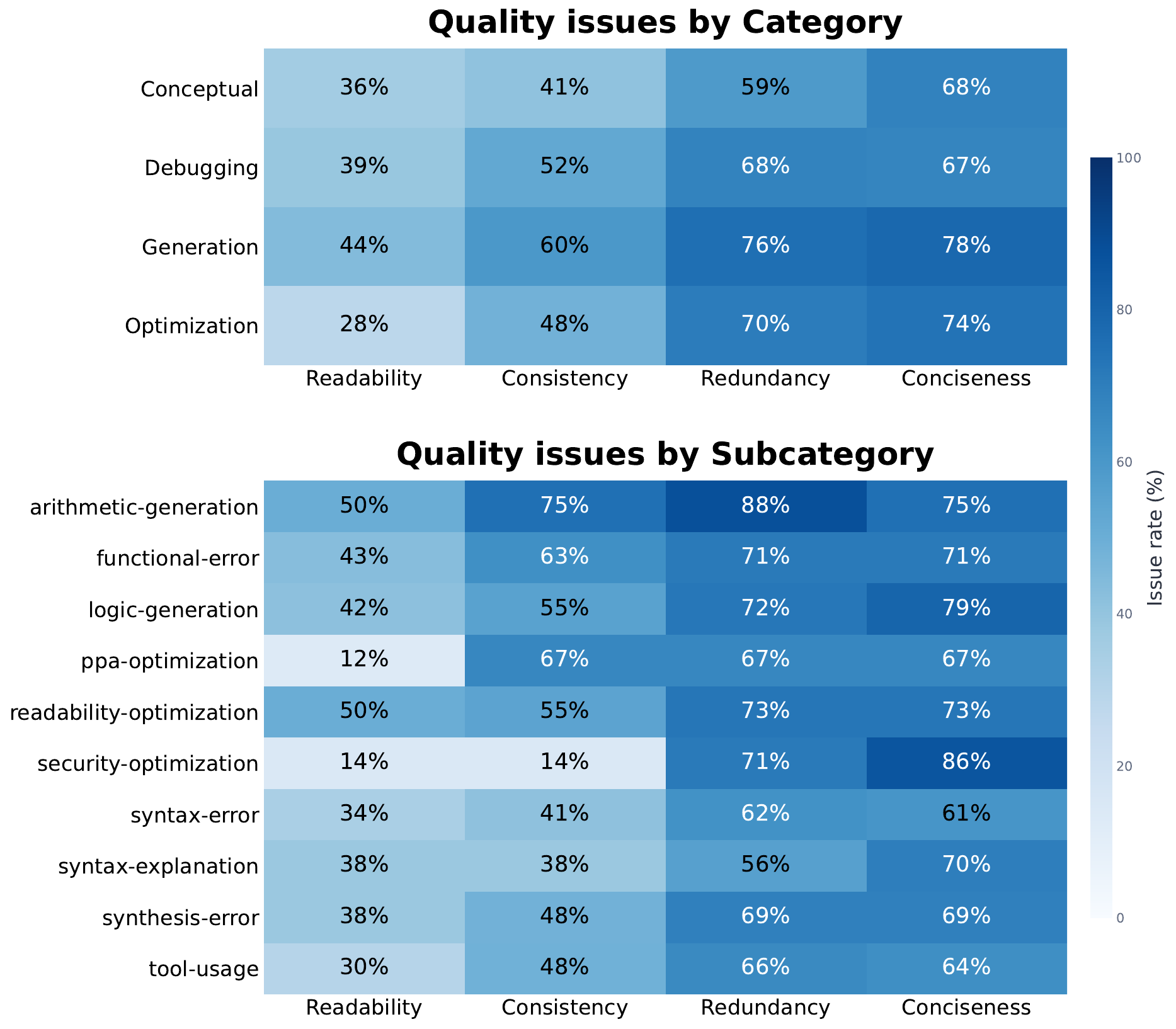} 
    \caption{Task-dependent severity of LLM answer quality issues.}
    \label{fig:task_heatmap}
    \vspace{-1em}
\end{figure}

LLM answer quality is strongly task-dependent (Fig.~\ref{fig:task_heatmap}): redundancy is most severe in generation and debugging tasks, generation also shows the highest verbosity, while conceptual tasks are comparatively lower. 
\begin{figure}[h]
    \centering
    \includegraphics[width=1\columnwidth]{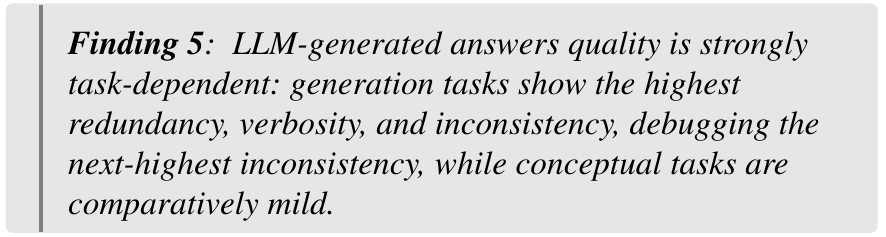} 
    \label{fig:task_severity}
    \vspace{-2em}
\end{figure}

\section{Multi-agent framework}
\label{sec:framework}

\begin{figure*}[!t]
    \centering
    \includegraphics[width=2\columnwidth]{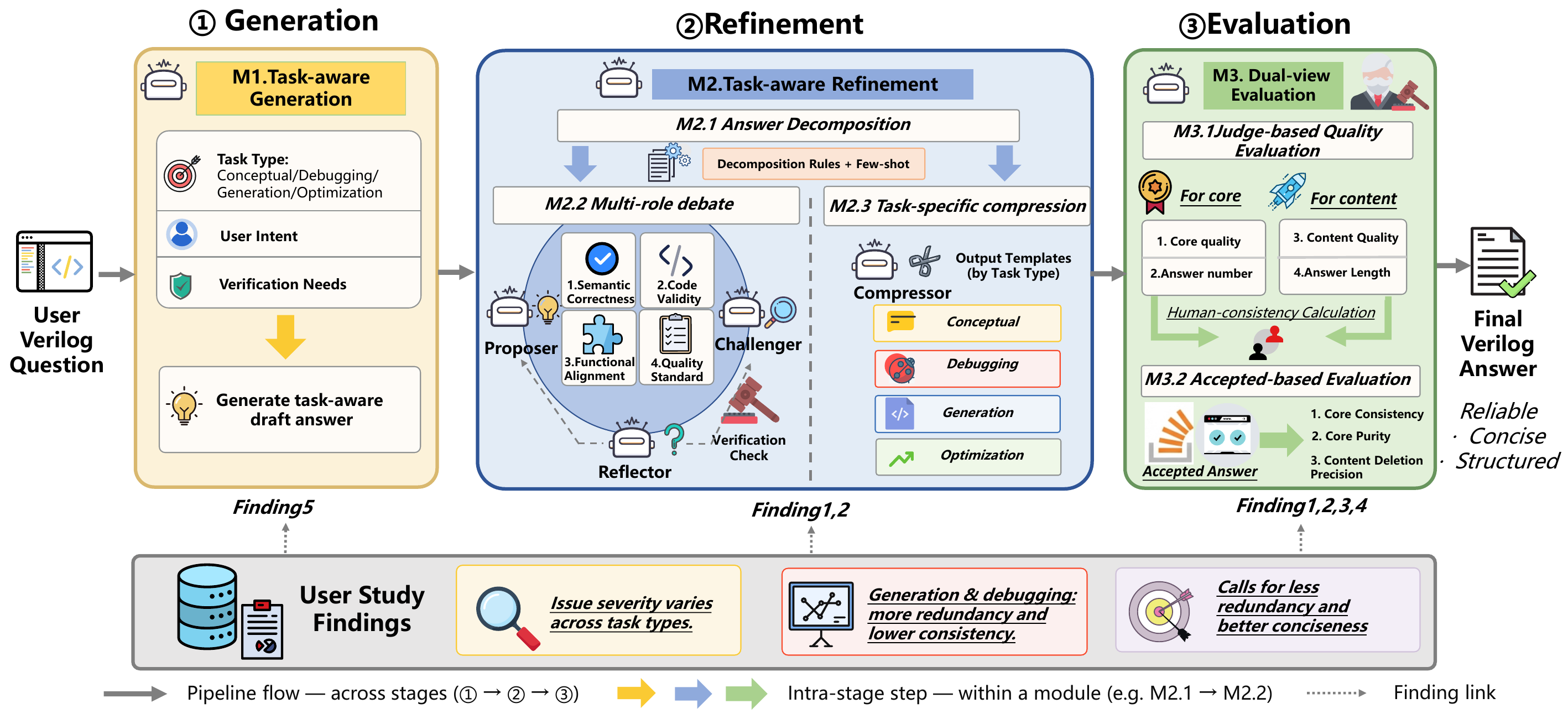} 
    \caption{Pipeline of the multi-agent framework. \colorbox{myyellow}{$M_{1}$} profiles the question (task type, intent, verification needs) and drafts an answer; \colorbox{myblue}{$M_{2}$} decomposes it into core and non-core content ($M_{2.1}$), prunes redundant core candidates via multi-role debate ($M_{2.2}$), and compresses non-core content by task type ($M_{2.3}$); \colorbox{mygreen}{$M_{3}$} scores the refined answer with an LLM-as-Judge ($M_{3.1}$) and validates it against the accepted answer ($M_{3.2}$). Each stage is driven by the corresponding user-study findings.}
    \label{fig:overview}
    \vspace{-1em}
\end{figure*}

\subsection{Overview}
We propose an LLM-based multi-agent framework that generates reliable, concise, and structured HDL answers through a task-aware division of labor: it first infers what kind of answer a question needs, then prunes the redundancy and verbosity LLMs tend to introduce while preserving task-critical content. As illustrated in Fig.~\ref{fig:overview}, given a user question, the framework operates in three sequential stages:

\begin{itemize}
\item \textbf{Task-aware Generation ($M_1$).} A Generation agent infers the question's task type, intent, and verification needs, then produces a task-aware draft.
\item \textbf{Task-aware Refinement ($M_2$).} The draft is split into core and non-core content; a multi-role debate prunes redundant core candidates while a task-specific compressor trims the non-core part.
\item \textbf{Dual-view Evaluation ($M_3$).} An LLM-as-Judge scores answer quality, and an accepted-based check validates the result against the human expert's answer.
\end{itemize}

This design directly responds to our findings (Section~\ref{sec:user_study}): since LLM answer quality is strongly task-dependent (Finding~5), a single uniform strategy is insufficient; profiling the task up front lets each stage target its characteristic failure mode---core redundancy (Finding~1) and non-core verbosity (Finding~2)---while preserving consistency and readability (Findings~3--4). Algorithm~\ref{alg:framework} summarizes the full procedure; each stage is color-coded to the user-study finding that motivates it.

\subsection{Task-aware Generation (\texorpdfstring{$M_1$}{M1})}
Before answering, $M_1$ reports a task-aware profile: the task type $\hat{t} \in \mathcal{T} = \{\text{Conceptual}, \text{Debugging}, \text{Generation}, \text{Optimization}\}$, the user intent $u$, and the verification needs $v$ (e.g., syntax correctness, synthesizability, or functional alignment). It then generates a draft $a_0 = M_1(q, \hat{t}, u, v)$. The task type shapes every later stage---it selects the generation strategy, defines what counts as a core answer during decomposition, and determines the compression policy---so that conceptual questions yield an explanation, debugging questions a fix, generation questions correct code, and optimization questions targeted suggestions. The full prompt and output schema for $M_1$ are in the supplementary material.


\subsection{Task-aware Refinement (\texorpdfstring{$M_2$}{M2})}

\subsubsection{Answer Decomposition ($M_{2.1}$)}
The draft is first decomposed into two streams:\begin{equation} D(a_0) = \big(\, \underbrace{\{c_1, \ldots, c_n\}}_{\text{core answers}},\ \underbrace{\{e_1, \ldots, e_m\}}_{\text{non-core content}} \,\big), \end{equation} where each core answer $c_i$ is an independent block that directly addresses the question (e.g., a distinct fix or alternative implementation), and each unit $e_j$ is non-core text such as background, problem restatement, or closing summary. Separating the two streams lets redundancy and verbosity be handled by distinct mechanisms---the debate and the compressor respectively.


\subsubsection{Multi-role Debate ($M_{2.2}$)}
The core candidates are examined \emph{jointly} by a panel of role-playing agents---a proposer, a challenger, and a reflector---together with a verification check over four aspects: semantic correctness, code validity, functional alignment, and answer-quality standards. The concrete keep/eliminate criteria are listed in the supplementary material. Judging the candidates together rather than in isolation is key: it eliminates not only invalid candidates but also redundant ones that are subsumed by, or derivative variants of, a stronger retained candidate, driving the answer toward a single decisive solution (Finding~1). The surviving candidates form the verified core $a_1$ ($|a_1| \le n$).


\subsubsection{Task-aware Compression ($M_{2.3}$)}
Each non-core unit $e_j$ is assigned one of three dispositions by a \emph{Compressor}, conditioned on the task type $\hat{t}$:\begin{equation}d_j = \pi_{\hat{t}}(e_j) \in \{\text{keep}, \text{compress}, \text{delete}\}.\end{equation}The policy \emph{keeps} task-critical content (e.g., design notes for generation, trade-off analysis for optimization), \emph{compresses} merely verbose explanation, and \emph{deletes} pure echo, problem restatement, and closing summaries. This three-way disposition, rather than uniform truncation, drives the non-core length reduction while preserving task-critical information, so it does not lower answer quality.
\begin{algorithm}[t]
\caption{Multi-agent HDL QA Refinement Framework}
\label{alg:framework}
\begin{algorithmic}[1]
\Require question $q$, accepted answer $a^*$
\Ensure refined answer $a_r$, scores $(Q_c, N, Q_n, L, \mathrm{CC}, \mathrm{CP}, \mathrm{DP})$

\State $\hat{t} \leftarrow \textsc{TaskProfiler}(q)$
  \Comment{Conceptual / Debugging / Generation / Optimization}
\State $a_0 \leftarrow M_1(q,\,\hat{t},\,\text{intent},\,\text{verification needs})$

\State $\{c_1\ldots c_n\}, \{e_1\ldots e_m\} \leftarrow \textsc{Decompose}(a_0, \hat{t})$
  \Comment{core answers / non-core units}
\For{each $c_i$}
  \State \textbf{Proposer} $\to$ \textbf{Challenger} $\to$ \textbf{Reflector} $\to$ verdict $\in$ \{keep, eliminate\}
\EndFor
\State $a_1 \leftarrow \{c_i \mid \text{verdict} = \text{keep}\}$

\For{each $e_j$}
  \State $d_j \leftarrow \pi_{\hat{t}}(e_j) \in \{\text{keep, compress, delete}\}$
\EndFor
\State $a_r \leftarrow a_1 \cup \{\phi(e_j) \mid d_j \neq \text{delete}\}$

\State $(Q_c, N, Q_n, L) \leftarrow \textsc{LLMJudge}(a_r)$
\State $(\mathrm{CC}, \mathrm{CP}, \mathrm{DP}) \leftarrow \textsc{AcceptedEval}(a_r, a^*)$
\State \Return $a_r$, $(Q_c, N, Q_n, L, \mathrm{CC}, \mathrm{CP}, \mathrm{DP})$
\end{algorithmic}
\end{algorithm}

\subsection{Evaluation (\texorpdfstring{$M_3$}{M3}): Dual-View Evaluation}

\subsubsection{Judge-based Quality Evaluation ($M_{3.1}$)}$M_{3.1}$ scores the answer along its two parts:\begin{equation}M_{3.1}(a)=(Q_c,\,N,\,Q_n,\,L),\end{equation}where $Q_c \in [1,5]$ and $N$ capture core quality and redundancy, and $Q_n \in [1,5]$ and $L$ capture non-core quality and verbosity. All four are computed for both the unrefined and refined answers, allowing the effect of refinement to be measured directly; $N$ and $L$ correspond to ``Core Number'' and ``Content Length'' in Table~\ref{tab:main_results_core_noncore}.


\subsubsection{Accepted-based Evaluation ($M_{3.2}$)}
Quality scores alone do not verify whether refinement selects the right core or deletes only dispensable content, so $M_{3.2}$ compares the framework's decisions against the accepted answer. With $C$, $C^{*}$ the model-selected and accepted core and $D$ the deleted non-core units,
\begin{equation}
\mathrm{CC}=\frac{|C\cap C^{*}|}{|C^{*}|},\quad
\mathrm{CP}=\frac{|C\cap C^{*}|}{|C|},\quad
\mathrm{DP}=\frac{|D\setminus C^{*}|}{|D|}.
\end{equation}
Core Consistency (CC) and Core Purity (CP) measure the overlap of the model-selected core with the accepted core (relative to the accepted core and to the model core respectively), and Deletion Precision (DP) the fraction of deleted units absent from the accepted answer.

\section{Evaluation}

\subsection{Experimental Setup}

\subsubsection{Evaluation Dataset}
We evaluate our framework on the 363-question stratified sample used in the user study (Section~\ref{sec:user_study}), which preserves the category distribution of the full 6{,}246-question dataset across the four task types.

\subsubsection{Compared Methods}
We compare two settings that share the same backbone LLMs and differ only in how the answer is produced.
\textbf{Only LLM.} The model receives the raw question (title and body, including any embedded code) and answers in a zero-shot setting. 
\textbf{Ours.} The model receives the same raw question, but the answer is generated and refined through our task-aware multi-agent framework ($M_1$ generation, $M_2$ refinement, $M_3$ judge). 
We have defined the maximum number of rounds for the maximum debate as 3~\cite{du2024debate}; if no agreement is reached, the final decision will be made by the Reflector. Full models and hyperparameters are in the supplementary material.

\begin{table*}[h]
\centering
\caption{Main Results of Core and Non-core Answer Refinement}
\label{tab:main_results_core_noncore}
\renewcommand{\arraystretch}{1.15}
\setlength{\tabcolsep}{4.2pt}
\scriptsize
\resizebox{\textwidth}{!}{
\begin{tabular}{llccc ccc ccc ccc}
\toprule
\multirow{3}{*}{\textbf{Model}} 
& \multirow{3}{*}{\textbf{Category}}
& \multicolumn{6}{c}{\textbf{Core Answer Refinement}}
& \multicolumn{6}{c}{\textbf{Non-core Content Refinement}} \\
\cmidrule(lr){3-8} \cmidrule(lr){9-14}
& 
& \multicolumn{3}{c}{\textbf{Core Quality} $\uparrow$}
& \multicolumn{3}{c}{\textbf{Core Number} $\downarrow$}
& \multicolumn{3}{c}{\textbf{Content Quality} $\uparrow$}
& \multicolumn{3}{c}{\textbf{Content Length} $\downarrow$} \\
\cmidrule(lr){3-5} \cmidrule(lr){6-8}
\cmidrule(lr){9-11} \cmidrule(lr){12-14}
& 
& LLM & Ours & $\Delta$
& LLM & Ours & Red.
& LLM & Ours & $\Delta$
& LLM & Ours & Red. \\
\midrule

\rowcolor{gray!20}
\cellcolor{white} & Debugging & 3.52 & 4.82 & +1.30 & 3.23 & 1.80 & 44\% & 3.92 & 4.28 & +0.36 & 1275 & 940 & 26\% \\
 & Conceptual & 3.68 & 4.41 & +0.73 & 3.45 & 2.27 & 34\% & 3.90 & 4.32 & +0.42 & 777 & 474 & 39\% \\
\rowcolor{gray!20}
\cellcolor{white}Qwen3.5 & Generation & 3.00 & 4.50 & +1.50 & 4.75 & 3.75 & 21\% & 3.97 & 4.13 & +0.16 & 1172 & 1036 & 12\% \\
 & Optimization & 2.50 & 4.17 & +1.67 & 2.33 & 1.33 & 43\% & 4.00 & 4.59 & +0.59 & 1571 & 1455 & 7\% \\
\rowcolor{gray!20}
\cellcolor{white} & \textbf{All} & \textbf{3.46} & \textbf{4.63} & \textbf{+1.17} & \textbf{3.30} & \textbf{2.00} & \textbf{39\%} & \textbf{3.92} & \textbf{4.29} & \textbf{+0.37} & \textbf{1108} & \textbf{815} & \textbf{26\%} \\

\midrule

 & Debugging & 3.69 & 4.65 & +0.96 & 3.96 & 2.38 & 40\% & 3.77 & 4.27 & +0.50 & 956 & 666 & 30\% \\
\rowcolor{gray!20}
\cellcolor{white} & Conceptual & 3.97 & 4.75 & +0.78 & 3.88 & 2.68 & 31\% & 3.74 & 4.22 & +0.48 & 543 & 236 & 57\% \\
DeepSeek-V4 & Generation & 3.20 & 4.80 & +1.60 & 3.40 & 1.80 & 47\% & 3.54 & 4.00 & +0.46 & 1033 & 906 & 12\% \\
\rowcolor{gray!20}
\cellcolor{white} & Optimization & 3.88 & 4.10 & +0.22 & 4.33 & 2.67 & 38\% & 3.82 & 3.91 & +0.09 & 707 & 470 & 33\% \\
 & \textbf{All} & \textbf{3.65} & \textbf{4.63} & \textbf{+0.98} & \textbf{3.95} & \textbf{2.46} & \textbf{38\%} & \textbf{3.74} & \textbf{4.21} & \textbf{+0.47} & \textbf{805} & \textbf{526} & \textbf{35\%} \\

\midrule

\rowcolor{gray!20}
\cellcolor{white} & Debugging & 3.84 & 4.80 & +0.96 & 5.65 & 3.64 & 36\% & 3.66 & 4.05 & +0.39 & 1299 & 1030 & 21\% \\
 & Conceptual & 4.15 & 4.91 & +0.76 & 5.32 & 3.72 & 30\% & 3.74 & 4.03 & +0.29 & 876 & 451 & 49\% \\
\rowcolor{gray!20}
\cellcolor{white}GPT-5.4 & Generation & 3.88 & 4.75 & +0.87 & 5.44 & 3.19 & 41\% & 3.83 & 4.25 & +0.42 & 1243 & 954 & 23\% \\
 & Optimization & 3.83 & 4.83 & +1.00 & 5.17 & 3.33 & 36\% & 3.55 & 3.82 & +0.27 & 1409 & 1183 & 16\% \\
\rowcolor{gray!20}
\cellcolor{white} & \textbf{All} & \textbf{3.94} & \textbf{4.83} & \textbf{+0.89} & \textbf{5.52} & \textbf{3.63} & \textbf{34\%} & \textbf{3.70} & \textbf{4.05} & \textbf{+0.35} & \textbf{1137} & \textbf{806} & \textbf{29\%} \\

\midrule

 & Debugging & 3.85 & 4.46 & +0.61 & 3.08 & 1.81 & 41\% & 3.55 & 4.42 & +0.87 & 1200 & 772 & 36\% \\
\rowcolor{gray!20}
\cellcolor{white} & Conceptual & 3.71 & 4.65 & +0.94 & 3.53 & 2.12 & 40\% & 3.50 & 4.40 & +0.90 & 927 & 451 & 51\% \\
Gemini-3.1 & Generation & 4.00 & 5.00 & +1.00 & 2.50 & 1.50 & 40\% & 3.60 & 4.20 & +0.60 & 1332 & 1191 & 11\% \\
\rowcolor{gray!20}
\cellcolor{white} & Optimization & 3.75 & 4.50 & +0.75 & 3.75 & 2.75 & 27\% & 3.44 & 4.11 & +0.67 & 1234 & 1072 & 13\% \\
 & \textbf{All} & \textbf{3.80} & \textbf{4.57} & \textbf{+0.77} & \textbf{3.24} & \textbf{1.96} & \textbf{40\%} & \textbf{3.53} & \textbf{4.38} & \textbf{+0.85} & \textbf{1121} & \textbf{720} & \textbf{36\%} \\
\midrule
\rowcolor{green!10}
\cellcolor{white} \textbf{Average} & \textbf{All} & \textbf{3.71} & \textbf{4.67} & \textbf{+0.96} & \textbf{4.00} & \textbf{2.51} & \textbf{37\%} & \textbf{3.72} & \textbf{4.23} & \textbf{+0.51} & \textbf{1043} & \textbf{717} & \textbf{31\%} \\
\bottomrule

\end{tabular}
}
\end{table*}

\subsection{Judge-based Quality Evaluation}
\subsubsection{Overall results}
Table~\ref{tab:main_results_core_noncore} reports the results across the four LLMs and four task categories. Our framework improves quality on both parts of the answer while substantially reducing core answer number and non-core content length. On average, core quality rises from 3.71 to 4.67 (+0.96) and content quality from 3.72 to 4.23 (+0.51), while the number of core answers drops by 37\% (4.00$\to$2.51) and non-core content length by 31\% (1043$\to$717). Crucially, the reductions in core answer number and non-core content length do not come at the cost of quality; both quality scores instead increase, indicating that the removed content was largely redundant or non-essential rather than informative.

\subsubsection{Consistency across models}
The improvement holds for all four backbones: post-refinement core quality is consistently high (4.57--4.83) from noticeably different baselines (3.46--3.94), and the gain is largest on the weakest baseline (Qwen3.5, +1.17), showing the framework is largely model-agnostic and converges models to a similar level. Redundancy falls consistently as well (Table~\ref{tab:main_results_core_noncore}).

\subsubsection{Task-dependent effects}
Refinement varies across task types in an interpretable way. Generation tasks see the largest core-quality gains but the smallest length reductions, because their answers are dominated by load-bearing code whose rationale is hard to compress; the gain there comes mainly from removing redundant alternative implementations. Conceptual tasks show the opposite---large length reductions but more modest quality gains---since they carry substantial removable background while their core explanation is already focused (Table~\ref{tab:main_results_core_noncore}).

\subsubsection{Reliability of the LLM-as-Judge}
To verify that $M_3$ is a reliable evaluator, we compare its scores against human annotators on 40 sampled answers. On both core and non-core quality, $M_3$ matches or exceeds human--human agreement (core: MAE 0.113, Spearman 0.939, adjacent 100\%; non-core: MAE 0.131, Spearman 0.949, adjacent 100\%), supporting its use as the primary quality metric. Full per-metric results are reported in the supplementary material.

\subsection{Accepted-based Evaluation}

We next validate core selection against the \emph{accepted answers provided by human experts}---the answer chosen by the engineer who posed the question---as a naturally occurring reference. We assess two aspects: whether the model selects the same core answer the expert did, and whether the content it deletes is genuinely non-essential.

\subsubsection{Consistency with the human expert on core answer selection}
We measure the overlap between the model-selected core and the core of the \emph{accepted answer}---the answer chosen by the engineer who posed the question. Table~\ref{tab:core_consistency_purity} reports \emph{Core Consistency} (agreement between the two cores) and \emph{Core Purity} (the fraction of the model-selected core the accepted answer also treats as core), before and after refinement. Refinement raises both: consistency from 43.2\% to 48.1\% (Qwen3.5) and 56.5\% to 59.4\% (DeepSeek-V4), purity from 50.0\% to 55.3\% and 61.9\% to 65.5\%. The upward shift shows pruning moves the core toward the human expert, not away from it. The moderate absolute level ($\sim$50--60\%) is expected rather than a deficiency: a single accepted answer reflects one engineer's choice, while the model often reaches a different but equally valid solution path---so low overlap reflects the diversity of valid HDL answers rather than incorrect selection.

\begin{table}[t]   
\centering
\small             
\renewcommand{\arraystretch}{1.15}
\setlength{\tabcolsep}{4.5pt}  
\caption{Results of Core Consistency and Core Purity}
\label{tab:core_consistency_purity}
\begin{tabular}{llcc}
\toprule
\textbf{Model} & \textbf{Category} & \textbf{CC} & \textbf{CP} \\
\midrule
\rowcolor{gray!20}
\rowcolor{gray!20}
\cellcolor{white} & Debugging & 41.2\%$\rightarrow$45.0\% & 51.2\%$\rightarrow$58.1\% \\
 & Conceptual & 42.9\%$\rightarrow$54.3\% & 51.4\%$\rightarrow$54.3\% \\
\rowcolor{gray!20}
\cellcolor{white}Qwen3.5 & Generation & 70.0\%$\rightarrow$66.7\% & 42.4\%$\rightarrow$45.5\% \\
 & Optimization & 28.6\%$\rightarrow$28.6\% & 33.3\%$\rightarrow$33.3\% \\
\rowcolor{gray!20}
\cellcolor{white} & \textbf{All} & \textbf{43.2\%$\rightarrow$48.1\%} & \textbf{50.0\%$\rightarrow$55.3\%} \\

\midrule

 & Debugging & 52.8\%$\rightarrow$55.6\% & 52.8\%$\rightarrow$57.7\% \\
\rowcolor{gray!20}
\cellcolor{white} & Conceptual & 66.7\%$\rightarrow$70.4\% & 76.9\%$\rightarrow$82.0\% \\
DeepSeek-V4 & Generation & 50.0\%$\rightarrow$50.0\% & 83.3\%$\rightarrow$75.0\% \\
\rowcolor{gray!20}
\cellcolor{white} & Optimization & 25.0\%$\rightarrow$25.0\% & 25.0\%$\rightarrow$25.0\% \\
 & \textbf{All} & \textbf{56.5\%$\rightarrow$59.4\%} & \textbf{61.9\%$\rightarrow$65.5\%} \\
\bottomrule
\end{tabular}
\vspace{-1em}
\end{table}

\subsubsection{Precision of Non-core Deletion}
We further ask whether the removed content is genuinely non-essential, using the accepted answers provided by human experts as reference. Table~\ref{tab:context_deletion_precision} reports \emph{Deletion Precision}: the fraction of deleted units whose information is also absent from the non-core part of the accepted answer. Unlike core selection, this is robust to solution diversity, as it only checks whether removed material is absent from the reference. Precision is high across models and categories (97.4\% for Qwen3.5, 98.7\% for DeepSeek-V4, several reaching 100\%), showing the verbosity reduction in Table~\ref{tab:main_results_core_noncore} comes from discarding content the reference also omits.

\begin{table}[t]
\centering
\caption{Results of Non-core Content Deletion Precision}
\label{tab:context_deletion_precision}
\renewcommand{\arraystretch}{1.15}
\setlength{\tabcolsep}{8pt}
\begin{tabular}{llc}
\toprule
\textbf{Model} & \textbf{Category} & \textbf{DP} \\
\midrule

\rowcolor{gray!20}
\cellcolor{white} & Debugging & 95.8\% \\
& Conceptual & 100.0\% \\
\rowcolor{gray!20}
\cellcolor{white}Qwen3.5 & Generation & 95.2\% \\
 & Optimization & 90.0\% \\
\rowcolor{gray!20}
\cellcolor{white} & \textbf{All} & \textbf{97.4\%} \\

\midrule

 & Debugging & 96.5\% \\
\rowcolor{gray!20}
\cellcolor{white} & Conceptual & 100.0\% \\
DeepSeek-V4 & Generation & 100.0\% \\
\rowcolor{gray!20}
\cellcolor{white} & Optimization & 100.0\% \\
 & \textbf{All} & \textbf{98.7\%} \\

\bottomrule
\end{tabular}
\end{table}

\subsection{Case Study}
We illustrate the refinement with one representative debugging case drawn directly from the evaluation set. It shows that the redundancy reductions come from removing redundant core answers.

\begin{figure}[!t]
    \centering
    \includegraphics[width=1\columnwidth]{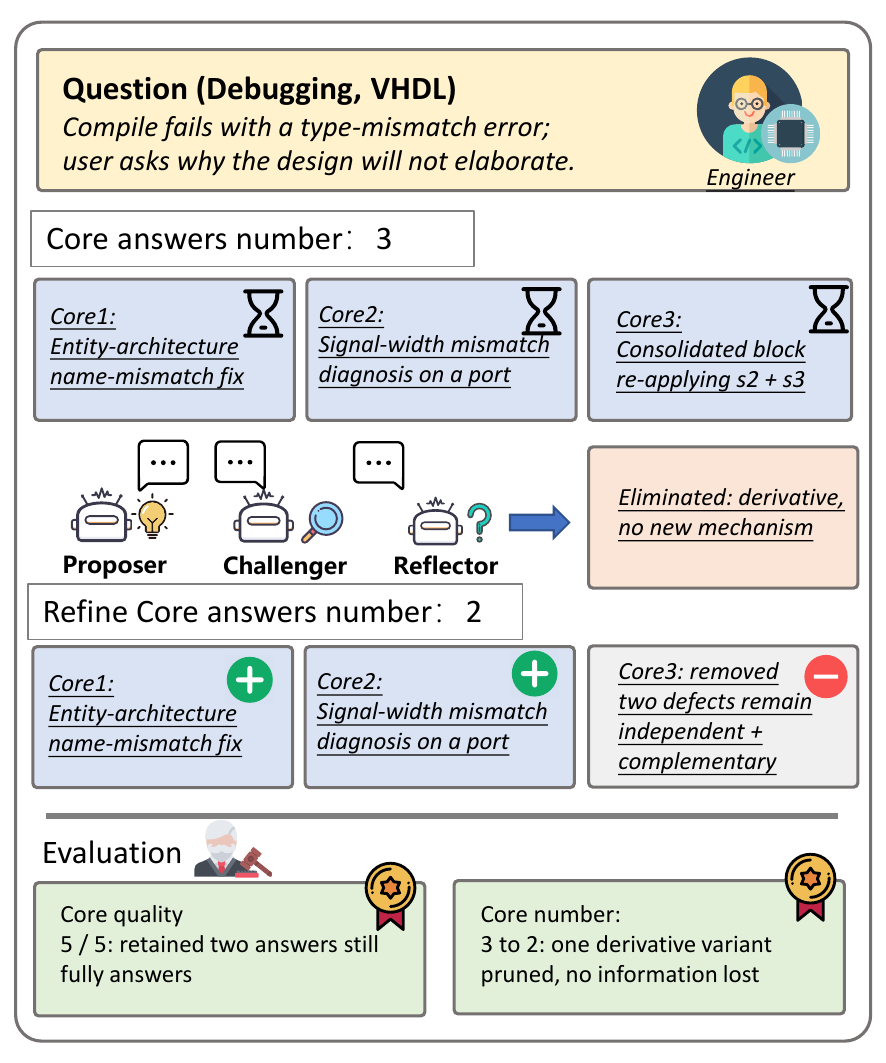} 
    \caption{Case study showing how multi-role debate prunes a redundant debugging solution without removing substantive information.}
    \label{fig:case1}
    \vspace{-2em}
\end{figure}

For a VHDL compile-error question, the draft was decomposed into three core candidates: (Core1) the entity--architecture name-mismatch fix, (Core2) the signal-width-mismatch diagnosis, and (Core3) a consolidated corrected-code block. The multi-role debate ($M_{2.2}$) kept Core1 and Core2 as two independent, complementary defects, but eliminated Core3 on the grounds that it merely re-packaged the fixes already contained in Core1 and Core2 without introducing a materially different mechanism---a derivative variant rather than an independent solution. This reduced the core-answer count from three to two. Crucially, the elimination removed no information: the LLM-as-Judge scored the refined core $5/5$, confirming that the retained two-candidate core still fully answers the question. This case shows how the \#Answers reduction in Table~\ref{tab:main_results_core_noncore} arises from discarding overlapping solutions, not correct answer. A second case study on a conceptual question is in the supplementary material.



\section{Discussion}

\subsection{Does a Single Prompt Suffice?}
A natural objection is whether a single prompt can match the framework. We compare against a strong \emph{single-prompt} setting that is explicitly instructed to remove redundancy while preserving solution diversity, evaluated with the same judge and metrics on the subset of questions with an accepted answer. As Table~\ref{tab:single-prompt} shows, on both backbones our framework attains higher core consistency and core purity (e.g., $+10.6$ and $+7.2$ points on Qwen3.5; $+5.1$ and $+8.4$ on DeepSeek-V4), even though the single prompt carries no redundancy-removal burden of its own. These results indicate that the gains stem from the multi-agent decomposition and debate rather than from the refinement objective alone: stating the goal in one prompt is not equivalent to enforcing it through structured selection.

\begin{table}[t]
\centering
\caption{Core Consistency (CC) and Core Purity (CP): a single redundancy-removal prompt vs.\ our frameworks.}
\label{tab:single-prompt}
\setlength{\tabcolsep}{6pt}
\renewcommand{\arraystretch}{1.15}
\begin{tabular}{llcc}
\toprule
\textbf{Model} & \textbf{Method} & \textbf{CC} & \textbf{CP} \\
\midrule

\multirow{2}{*}{Qwen3.5}
&\cellcolor{gray!20}  Single-Prompt &\cellcolor{gray!20}  37.5\% &\cellcolor{gray!20} 48.1\% \\

 & Ours          & \textbf{48.1\%} & \textbf{55.3\%} \\
\midrule
\multirow{2}{*}{DeepSeek-V4}
 &\cellcolor{gray!20} Single-Prompt &\cellcolor{gray!20}54.3\% &\cellcolor{gray!20} 57.1\% \\
 & Ours          & \textbf{59.4\%} & \textbf{65.5\%} \\
\bottomrule
\end{tabular}
\end{table}

\subsection{Threats to Validity}
\subsubsection{Internal Validity}
Our quality evaluation relies on an LLM-as-Judge, which may favor concise, well-structured outputs and prefer outputs stylistically similar to its own; we mitigate this by validating it against human annotators, whose agreement it matches. More fundamentally, we do not perform tool-level correctness verification, because a large fraction of both LLM-generated and accepted answers are code \emph{fragments} or natural-language reasoning that do not form self-contained, synthesizable units. We instead use consistency with the accepted answer (CC/CP), but stress that this measures \emph{non-divergence from a human reference, not correctness per se}. Our stronger, diversity-robust guarantee is on the non-core side, via Deletion Precision (97--98\%).

The task profiler $M_1$ is imperfect, so questions may be misclassified; the framework is largely robust to this, since the debate ($M_{2.2}$) and accepted-based evaluation ($M_{3.2}$) operate on answer content rather than task labels, and gains are consistent across categories (Table~\ref{tab:main_results_core_noncore}), though residual effects on the type-conditioned compression policy remain possible. In the per-model comparison, each question is answered by one randomly assigned model, so the backbones are evaluated on disjoint subsets, and cross-model differences may partly reflect question difficulty rather than capability. Most questions are also single-annotated---justified for the relatively objective redundancy and conciseness, but leaving more residual uncertainty on consistency and readability.

\subsubsection{External Validity}
The Stack Overflow snapshot predates the evaluated models, so potential training-data contamination could inflate apparent agreement. The user study involves 19 engineers with 1--3 years of experience; more senior engineers may weigh ``core'' versus ``redundant'' content differently. The dataset is a fixed-date snapshot, and the evaluation targets four specific models under default decoding; since redundancy and verbosity are sensitive to model and sampling choices, absolute rates may differ elsewhere, even if the qualitative gap persists.

\subsection{Practical implications}

As EDA tooling increasingly surfaces LLM answers within the design loop, a refinement layer of this kind offers a concrete reduction in the reading burden placed on engineers---a 37\% drop in answer count and a 31\% drop in non-core length---while preserving answer quality. This suggests that the value of an LLM assistant for hardware design depends not only on whether it can produce a correct answer, but on whether it can present that answer concisely.

Beyond the framework itself, the released dataset of 6,246 expert-validated HDL Q\&A pairs serves as a resource for retrieval-augmented generation pipelines, enabling LLMs to ground HDL responses in peer-validated engineering practice rather than parametric knowledge alone. The structured task taxonomy and accepted-answer annotations further make the dataset suitable for fine-tuning or supervised alignment of LLMs on domain-specific HDL tasks, where category-aware training signals may reduce the over-answering tendency at the model level. Finally, the dataset---paired with the dual-view evaluation protocol---provides a reusable benchmark for assessing the communicative quality of future HDL tools and models, complementing existing functional-correctness benchmarks such as VerilogEval and RTLLM.

\section{Related work}
\subsection{LLMs for HDL}
The application of LLMs to HDL has grown rapidly since the first systematic study of LLM-based RTL generation~\cite{Benchmarking}.
Subsequent work established benchmarks for functional correctness on specification-to-RTL tasks~\cite{verilogeval,rtllm}, with the recent CVDP benchmark~\cite{cvdp} showing that even state-of-the-art models remain far from saturating realistic design and verification problems.
A growing line of research targets the broader RTL workflow, including code generation~\cite{codev,origen,verilogcoder} and debugging and repair~\cite{rtlfixer,veridebug}. Beyond hardware, the software engineering community has long studied what constitutes a high-quality answer on Q\&A platforms such as Stack Overflow, identifying factors such as conciseness, comprehensiveness, and well-presented code examples as key determinants of answer usefulness and acceptance~\cite{nasehi2012,calefato2015}. This perspective has recently been extended to LLM-generated answers, with empirical studies comparing ChatGPT and human answers on Stack Overflow along dimensions including correctness, consistency, and conciseness~\cite{kabir2024}. 
We bring this answer-quality perspective into the HDL domain, studying the expert--LLM answer gap on real-world HDL questions. 

\subsection{Multi-agent frameworks}
Multi-agent frameworks coordinate several LLM agents to solve tasks beyond the reliability of a single model. Role-based cooperation assigns complementary roles through an explicit division of labor~\cite{metagpt,agentverse}, while multi-agent debate has agents iteratively critique and revise one another's outputs to improve factuality, consistency, and summarization quality~\cite{du2024debate,chateval}. 
Such systems have also been adapted to RTL design~\cite{verimind,rtlsquad}. Building on the role-based paradigm, we design a framework specialized for answer refinement with quality evaluation by an LLM-as-Judge.

\subsection{LLM-based Evaluation}

Reference-based metrics such as BLEU or ROUGE correlate poorly with human judgment on open-ended generation, as they reward surface n-gram overlap and penalize valid paraphrases or alternative solutions---a poor fit for HDL answers, where the same fix can be expressed in many equally correct forms. Recent work instead employs strong LLMs as evaluators, reporting agreement with humans comparable to human--human agreement~\cite{zheng2023judging,liu2023geval}, and such judges have been applied to dialogue, summarization, and code-related generation. 
In contrast, we apply an LLM judge to the core/non-core structure of HDL answers---separately assessing core technical quality and non-core conciseness---and validate its reliability against human annotators, finding agreement that matches or exceeds human--human agreement.
\section{Conclusion}

As engineers increasingly turn to LLMs for HDL question answering, answer quality---beyond mere fluency---directly affects how much effort they save. From a dataset of 6{,}246 real-world Stack Overflow HDL Q\&A pairs with accepted answers, a user study with 19 engineers quantified the gap: a pervasive \emph{over-answering} tendency in which correct content is buried under redundant alternatives and verbose padding. Motivated by these findings, we presented a task-aware multi-agent framework in which a generation agent ($M_1$) profiles each question, a multi-role debate ($M_2$) separates core from non-core content and prunes redundant material, and an evaluation harness ($M_3$) scores quality and checks against the accepted answer. Across four backbone models and four task categories, refinement cuts answer count by 37\% and non-core length by 31\% while both quality scores rise rather than fall, letting engineers reach the answer they need with less effort and making LLMs a more practical aid in the hardware design loop.
\section*{Acknowledgment}

During the preparation of this work, the authors used a generative AI assistant to improve the clarity of the text and to help generate parts of the experiment scripts. All AI-assisted output was reviewed, tested, and edited by the authors, who take full responsibility for the content of this work.

\bibliographystyle{IEEEtran} 
\bibliography{ref}

\end{document}


\title{Supplementary Material for: \\ When LLMs Over-Answer: Measuring and Mitigating Quality Issues in LLM-Based Hardware Description Language Question Answering}

\maketitle

\section*{Overview}
This supplementary material provides additional details and results that support, but are not essential to, the claims in the main paper. It contains eight parts, presented in order: (i) the dataset task-type classification scheme --- taxonomy, boundary-disambiguation rules, prompt template, and validation --- that produces the prior task label later double-checked by $M_1$ (Section~\ref{supp:taxonomy}); (ii) the structure of the user-study questionnaires and the per-dimension annotation rules given to participants (Section~\ref{supp:questionnaire}); (iii) the per-dimension inter-annotator agreement underlying the single-annotation design of our user study (Section~\ref{supp:iaa}); (iv) the implementation and experimental configuration, including models, decoding parameters, and key hyperparameters (Section~\ref{supp:config}); (v) the operative prompts for each agent, organized to mirror the method structure ($M_1$, $M_{2.1}$--$M_{2.3}$, $M_{3.1}$--$M_{3.2}$) of the main paper (Section~\ref{supp:prompts}); (vi) the full language distribution of our 6{,}246-question dataset (Section~\ref{supp:lang}); (vii) the full per-metric reliability comparison between the LLM-as-Judge and human annotators (Section~\ref{supp:judge-reliability}); and (viii) a second qualitative case study illustrating how the framework removes off-topic and summary content on a conceptual question (Section~\ref{supp:case2}). All section, table, and figure references in the main paper that point to ``the supplementary material'' resolve here.

\section{Dataset Task-Type Classification}
\label{supp:taxonomy}
Before the task-aware generation stage ($M_1$), every collected Q\&A record is assigned a task type by a single LLM call. This classification, together with its one-sentence justification, is the prior that $M_1$ later double-checks rather than recomputes. The classifier is given only the title and (truncated) question body plus the answers, is run at temperature $0$ for determinism, and is required to return a bare JSON object with no markdown fences. Each item is independently re-tried up to five times; responses that cannot be parsed are flagged as \texttt{parse\_error} (and API failures as \texttt{api\_error}) rather than being silently bucketed into a real category, so error items are excluded from accuracy rather than corrupting the distribution.

\subsection{Taxonomy}
The taxonomy is a two-level scheme of four primary categories, each with two or three subcategories, defined by \emph{primary intent}:
\begin{itemize}
  \item \textbf{Optimization} --- improve an already-working design. Subtypes: \texttt{ppa-optimization} (timing/power/area/cost), \texttt{security-optimization} (race conditions, metastability, glitches, sim--synth mismatch, leakage, testability), \texttt{readability-optimization} (style, naming, modularity, verbosity).
  \item \textbf{Generation} --- build new hardware from a functional requirement when no substantial implementation exists yet. Subtypes: \texttt{logic-generation} (FSMs, counters, memories, FIFOs, arbiters, \ldots), \texttt{arithmetic-generation} (adders, multipliers, dividers, MACs, FP/DSP units).
  \item \textbf{Debugging} --- fix an explicit defect in existing, substantially complete code. Subtypes: \texttt{syntax-error} (compilation/elaboration), \texttt{functional-error} (compiles but wrong simulation result), \texttt{synthesis-error} (tool-reported synthesis/P\&R error or warning).
  \item \textbf{Conceptual} --- understand a language feature, syntax rule, or EDA tool, with no PPAC/security/reliability/readability concern. Subtypes: \texttt{syntax-explanation} (semantics and idioms), \texttt{tool-usage} (compiler/EDA operation).
\end{itemize}

\subsection{Boundary Disambiguation}
Because intent rather than surface form decides the label, the prompt supplies explicit tie-breakers for the frequently confused pairs: Conceptual vs.\ Optimization (any PPAC/reliability concern, even phrased as ``does X guarantee Y,'' goes to Optimization), Generation vs.\ Conceptual (a stated functional requirement with only a stub still counts as Generation), Conceptual vs.\ Debugging (a concrete defect to fix is Debugging regardless of code presence), Debugging vs.\ Optimization (broken vs.\ correct-but-suboptimal), Debugging vs.\ Generation (substantial existing code vs.\ starting from scratch), \texttt{functional-error} vs.\ \texttt{synthesis-error} (error surfaced in simulation vs.\ in synthesis/P\&R), and \texttt{security-optimization} vs.\ \texttt{synthesis-error} (design-level defect vs.\ a tool warning needing no design fix).

\subsection{Prompt Template}
The operative classifier prompt is reproduced below; \texttt{\{title\}}, \texttt{\{body\}}, \texttt{\{answers\}}, and \texttt{\{data\_id\}} are filled per item.
\begin{quote}\small\ttfamily
You are an expert hardware-design Q\&A classifier. Your task is to assign exactly one category and subcategory to the following Stack Overflow post about Verilog or hardware design. [INPUT DATA: id, title, body, answers] [TAXONOMY: the four categories and ten subcategories above, each with its intent definition] [BOUNDARY DISAMBIGUATION: the deciding-question table above] Return ONLY a valid JSON object --- no markdown fences, no extra text --- with fields id, category (Conceptual$|$Debugging$|$Generation$|$Optimization), subcategory (one of the ten), and reason (one sentence citing specific evidence from the post).
\end{quote}

\subsection{Validation}
On a held-out subset with human-labeled ground truth (category in column~A, subcategory in column~B of an Excel sheet, aligned by position), the pipeline tracks running category- and subcategory-level accuracy, reporting an overall figure plus a per-category breakdown; \texttt{parse\_error}/\texttt{api\_error} items are excluded from these figures. A separate full-dataset mode runs the identical classifier without ground truth to label the complete corpus.

\section{User-Study Questionnaire Structure and Per-Dimension Annotation Rules}
\label{supp:questionnaire}
This section documents the structure of the user-study questionnaire and the per-dimension rules given to annotators, complementing the inter-annotator agreement analysis in Section~\ref{supp:iaa}.

\subsection{Questionnaire Structure}
The full questionnaire is partitioned into 19 questionnaire distributed to annotators, each containing 19-20 questions. Each question pairs a real HDL engineering problem with two candidate answers (A and B), one of which is the engineer (Golden) answer and the other the LLM answer; annotators read both independently before scoring. Each item exposes four structured elements: a category tag (e.g., \texttt{Conceptual / syntax-explanation}, \texttt{Debugging}, \texttt{Generation}, \texttt{Optimization}), the question title, the question description (taken verbatim from the original Q\&A), and the two candidate answers. To remove position bias, the A/B presentation order is randomized per question; the true identity of each answer (which is Golden, which is LLM) is disclosed beneath the title, and annotators are instructed to judge on content quality alone. Every item is rated on four quality dimensions (an additional overall-preference field is collected but not used here).

\subsection{Per-Dimension Annotation Rules}
\paragraph{(1) Readability --- single choice.} Which answer is the better read. Options: A better / B better / tie / cannot judge. This is a \emph{comparative} judgment over A and B.

\paragraph{(2) Consistency --- single choice.} The degree to which the \emph{LLM answer} covers the core content of the \emph{engineer (Golden) answer}; the per-item identity labels fix the comparison direction. Options: fully covered / partially covered / missing / cannot judge.

\paragraph{(3) Redundancy --- judged for A and B separately.} Whether the \emph{core} of an answer contains repeated or superfluous content. Options (one set each for A and B): yes (redundant) / no (not redundant) / cannot judge. Category-specific guidance is provided: for Debugging, whether multiple solutions are given; for Conceptual, whether the same concept is explained multiple times; for Generation, whether multiple code blocks are given; for Optimization, whether multiple optimization suggestions are given.

\paragraph{(4) Conciseness --- judged for A and B separately.} Whether, beyond the core content and necessary supporting material (e.g., code examples), there is unnecessary content such as superfluous background, over-summarization, or irrelevant elaboration. Options (one set each for A and B): yes (not concise) / no (concise) / cannot judge.

\subsection{Design Rationale}
The dimensions deliberately differ in granularity: readability and consistency are \emph{comparative} single-choice judgments over A and B (consistency further binding the Golden$\rightarrow$LLM direction), whereas redundancy and conciseness are answered \emph{independently} for each answer. Redundancy targets repetition \emph{within} the core, while conciseness targets superfluous material \emph{outside} the core; the two are complementary --- the former asks how many times the necessary content is stated, the latter asks whether unnecessary content has crept in.

\section{Inter-Annotator Agreement}
\label{supp:iaa}
To assess the reliability of the single-annotation design used in the user study, a second annotator independently re-labeled a subset of questions, with both annotators judging all four quality dimensions on the same items. We report percent agreement and Cohen's $\kappa$~\cite{landis1977} for each dimension in Table~\ref{tab:iaa}. Redundancy and conciseness, which rest on relatively objective criteria, show the highest agreement ($\kappa=0.84$ and $0.89$), while the more subjective readability and consistency dimensions remain in the substantial range ($\kappa=0.79$ and $0.83$). The macro-average across dimensions is $91.5\%$ agreement and $\kappa=0.84$, which by the Landis--Koch scale~\cite{landis1977} corresponds to substantial-to-almost-perfect agreement and supports the single-annotation design.

\begin{table}[h]
\centering
\caption{Inter-annotator agreement on the doubly annotated subset.}
\label{tab:iaa}
\begin{tabular}{lcc}
\toprule
Dimension & \% Agreement & Cohen's $\kappa$ \\
\midrule
\rowcolor{gray!20}
Readability & 87.6 & 0.79 \\
Consistency & 90.7 & 0.83 \\
\rowcolor{gray!20}
Redundancy & 92.8 & 0.84 \\
Conciseness & 94.8 & 0.89 \\
\midrule
\rowcolor{gray!20}
Macro-average & 91.5 & 0.84 \\
\bottomrule
\end{tabular}
\end{table}

\section{Implementation and Experimental Configuration}
\label{supp:config}
Table~\ref{tab:config} lists the models, decoding parameters, and key hyperparameters used in all experiments. The framework is implemented on the AutoGen multi-agent library~\cite{wu2024autogen}. All stages query their backbone under temperature $0$ for determinism, so reported results are reproducible across re-runs. The multi-role debate ($M_{2.2}$) runs for at most three rounds; if no agreement is reached, the Reflector makes the final decision. Each stage can be routed to a different backbone, but in the reported configuration all generation and refinement stages share one model and the LLM-as-Judge ($M_{3.1}$) uses a separate model, to avoid a model scoring its own output.

\begin{table}[h]
\centering
\caption{Models and hyperparameters used in the experiments.}
\label{tab:config}
\renewcommand{\arraystretch}{1.2}
\setlength{\tabcolsep}{6pt}
\begin{tabular}{l >{\raggedright\arraybackslash}p{0.55\columnwidth}}
\toprule
\textbf{Setting} & \textbf{Value} \\
\midrule
\rowcolor{gray!20}
Backbone models ($M_1$--$M_2$) & Qwen3.5, DeepSeek-V4, GPT-5.4, Gemini-3.1 \\
LLM-as-Judge ($M_3$) & separate from the answer model \\
\rowcolor{gray!20}
Decoding temperature & 0.0 (deterministic) \\
Top-$p$ & provider default \\
\rowcolor{gray!20}
Max debate rounds ($M_{2.2}$) & 3 \\
Tie-break in debate & Reflector decides \\
\rowcolor{gray!20}
Decomposition ($M_{2.1}$, $M_{2.3}$) & rules + few-shot \\
Multi-agent library & AutoGen \\
\bottomrule
\end{tabular}
\end{table}

\section{Agent Prompts}
\label{supp:prompts}
This section provides the operative prompts for each agent, organized to mirror the method structure in the main paper ($M_1$; $M_2$ with $M_{2.1}$/$M_{2.2}$/$M_{2.3}$; $M_3$ with $M_{3.1}$/$M_{3.2}$). All agents are queried at temperature $0$ and instructed to return valid JSON without markdown fences.

\subsection{Task-aware Generation ($M_1$)}
$M_1$ double-checks the dataset's prior task label rather than reclassifying from scratch, and emits the task type, user intent, verification needs, and answer policy as a single JSON object. Its system instruction is:
\begin{quote}\small\ttfamily
You are a requirements reviewer for Verilog/SystemVerilog/VHDL engineering questions. The dataset ALREADY provides a prior task-type classification together with its reasoning. Your job is to DOUBLE-CHECK that prior classification --- NOT to reclassify from scratch: keep it if reasonable, override only when clearly wrong and justify the override. Besides the (possibly corrected) task type, also produce the verification needs and answer policy.
\end{quote}
The output schema includes \texttt{task\_type}, \texttt{secondary\_type}, \texttt{user\_intent}, \texttt{verification\_needs} (a subset of \{semantic\_correctness, syntax\_check, synthesizability\_check, functional\_alignment, optimization\_validity\}), and an \texttt{answer\_policy} (target length, whether to include code or a testbench, whether to list multiple options). Verification needs are guided per category: concept/principle reasoning $\rightarrow$ semantic\_correctness; any code produced or edited $\rightarrow$ syntax\_check + synthesizability\_check; behavioural/simulation aspects $\rightarrow$ functional\_alignment; an optimization claim $\rightarrow$ optimization\_validity.

\subsection{Answer Decomposition ($M_{2.1}$)}
$M_{2.1}$ splits the draft into ordered CORE and CONTEXT segments under one definition, with a recall-first policy:
\begin{quote}\small\ttfamily
Split a draft into ordered segments and label each as CORE or CONTEXT under ONE definition: CORE = a block that DIRECTLY and SPECIFICALLY answers the user's actual question; CONTEXT = generic guidance, closing summaries/recaps, background or problem restatement, and loosely-related elaboration. RECALL FIRST: emit one CORE per distinct point that addresses the problem (multiple fixes / alternatives / approaches $\rightarrow$ multiple cores); default to splitting when unsure. Sequential build steps of ONE artifact, or one explanation split into bullets, stay as ONE core. Always emit at least one CORE.
\end{quote}

\subsection{Multi-role Debate ($M_{2.2}$)}
$M_{2.2}$ judges all core candidates jointly via a proposer/challenger/reflector panel, with a verification check over four aspects (semantic correctness, code validity, functional alignment, answer-quality standards). A candidate is kept if technically correct, usable, on-topic, constraint-compliant, and independently valuable; it is eliminated only for one of six concrete reasons:
\begin{quote}\small\ttfamily
ELIMINATE a candidate ONLY for: (1) TECHNICAL ERROR --- wrong HDL semantics, broken/non-synthesizable code, or a diagnosis that contradicts the symptom; (2) INFERIOR/REDUNDANT/DERIVATIVE ALTERNATIVE --- wrapper-only, parameter-only, scope-only, or weaker duplicate of a kept candidate; (3) GENERIC GUIDANCE not tied to the user's specific code; (4) VIOLATES A USER HARD CONSTRAINT; (5) OFF-TOPIC EXTRA --- solves a different problem the user did not ask about; (6) SPECULATIVE LONG-SHOT among guessed causes. HARD GUARD: never apply (6) to confirmed independent defects visible in posted code. NEVER eliminate for FORM differences (terseness, wording, style, omitting explanation).
\end{quote}
Each candidate receives a keep/eliminate verdict with a one-sentence reason naming the category; the surviving candidates form the verified core $a_1$.

\subsection{Task-aware Compression ($M_{2.3}$)}
$M_{2.3}$ decomposes the non-core stream into typed blocks and applies a task-aware keep/compress/delete policy $\pi_{\hat t}$. Load-bearing content per category is protected (design notes and usage constraints for generation; problem and impact analysis for optimization are \texttt{keep}); closing summaries and pure restatements are \texttt{delete} across all categories; remaining context is \texttt{compress}. The compressor is instructed:
\begin{quote}\small\ttfamily
Remove water --- redundancy, filler, hedging, repeated restatements --- while preserving EVERY genuine information point. Do NOT delete information, add information, change technical claims, or touch code (keep code verbatim). The result must say the SAME things with fewer words.
\end{quote}

\subsection{Judge-based Quality Evaluation ($M_{3.1}$)}
$M_{3.1}$ scores core quality $Q_c\in[1,5]$, non-core content quality $Q_n\in[1,5]$, the number of independent core answers $N$, and the non-core length $L$, for both the unrefined and refined answers.

\subsection{Accepted-based Evaluation ($M_{3.2}$)}
$M_{3.2}$ compares the framework's core selection and deletions against the human accepted answer, computing Core Consistency, Core Purity, and Deletion Precision as defined in the main paper.

\section{Language Distribution of the Dataset}
\label{supp:lang}
Table~\ref{tab:lang_dist} reports the full language distribution of the 6{,}246-question dataset. Verilog/SystemVerilog and VHDL account for 53.4\% and 43.2\% of all 6,246 records respectively, with the remaining 3.4\% consisting of cross-family, other-HDL, or untagged posts; this confirms the dataset spans both dominant HDL families rather than being Verilog-specific.

\begin{table}[h]
\centering
\caption{Language distribution of the HDL Q\&A dataset.}
\label{tab:lang_dist}
\renewcommand{\arraystretch}{1.15}
\setlength{\tabcolsep}{6pt}
\begin{tabular}{llcc}
\toprule
\textbf{Family} & \textbf{Language} & \textbf{\# Instances} & \textbf{\%} \\
\midrule
\rowcolor{gray!20}
Verilog (3,338) & Verilog & 2,348 & 37.6 \\
\rowcolor{gray!20}
 & SystemVerilog + Verilog & 957 & 15.3 \\
\rowcolor{gray!20}
 & SystemVerilog & 18 & 0.3 \\
\rowcolor{gray!20}
 & Verilog + other HDL & 15 & 0.2 \\
\midrule
VHDL (2,698) & VHDL & 2,689 & 43.1 \\
 & VHDL + other HDL & 9 & 0.1 \\
\midrule
\rowcolor{gray!20}
Other (210) & Verilog + VHDL (cross) & 99 & 1.6 \\
\rowcolor{gray!20}
 & Cross-family mixed & 29 & 0.5 \\
\rowcolor{gray!20}
 & Chisel / other HDL only & 39 & 0.6 \\
\rowcolor{gray!20}
 & No language tag & 43 & 0.7 \\
\midrule
Total & & 6,246 & 100.0 \\
\bottomrule
\end{tabular}
\end{table}

\section{Reliability of the LLM-as-Judge}
\label{supp:judge-reliability}
To verify that the LLM-as-Judge ($M_3$) is a reliable evaluator, we sampled 40 answers and had human annotators score them independently, then compared $M_3$'s scores against the human scores separately for core quality (Table~\ref{tab:m3_core_quality_agreement}) and non-core content quality (Table~\ref{tab:m4_context_quality_agreement}). As a reference point, we also report human--human agreement on the same items, so that $M_3$--human agreement can be read against the level of agreement humans reach with one another.

Across both parts of the answer, $M_3$--human agreement matches or exceeds human--human agreement on every metric. For core quality, $M_3$ attains an MAE~\cite{mae} of 0.113 (vs.\ 0.200 human--human), a Spearman correlation~\cite{spearman1961proof} of 0.939 (vs.\ 0.828), 100\% adjacent agreement (vs.\ 97.5\%), and 92.5\% directional agreement (vs.\ 82.5\%). The same pattern holds for non-core content quality (MAE 0.131, Spearman 0.949, 100\% adjacent agreement). These results indicate that $M_3$ is at least as consistent with human judgment as human annotators are with each other, supporting its use as the primary quality metric in our evaluation.

\begin{table}[h]
\centering
\caption{Human agreement results for core quality.}
\label{tab:m3_core_quality_agreement}
\begin{tabular}{lcc}
\toprule
\textbf{Metric} & \textbf{Human--Human} & \textbf{LLM--Human} \\
\midrule
\rowcolor{gray!20}
MAE & 0.200 & 0.113 \\
Adjacent Agreement & 97.5\% & 100.0\% \\
\rowcolor{gray!20}
Spearman $\rho$ & 0.828 & 0.939 \\
Quadratic Weighted $\kappa$~\cite{cohen1968weightedkappa} & 0.828 & 0.895 \\
\rowcolor{gray!20}
Directional Agreement & 82.5\% & 92.5\% \\
\bottomrule
\end{tabular}
\end{table}

\begin{table}[h]
\centering
\caption{Human agreement results for non-core content quality.}
\label{tab:m4_context_quality_agreement}
\begin{tabular}{lcc}
\toprule
\textbf{Metric} & \textbf{Human--Human} & \textbf{LLM--Human} \\
\midrule
\rowcolor{gray!20}
MAE & 0.212 & 0.131 \\
Adjacent Agreement & 100.0\% & 100.0\% \\
\rowcolor{gray!20}
Spearman $\rho$ & 0.848 & 0.949 \\
Quadratic Weighted $\kappa$ & 0.862 & 0.887 \\
\rowcolor{gray!20}
Directional Agreement & 80\% & 87.5\% \\
\bottomrule
\end{tabular}
\end{table}

\section{Additional Case Study}
\label{supp:case2}
This section complements the debugging case study in the main paper with a second example on a conceptual question, showing that the framework removes off-topic and summary content---not only redundant alternative solutions.

\begin{figure}[h]
\centering
\includegraphics[width=\columnwidth]{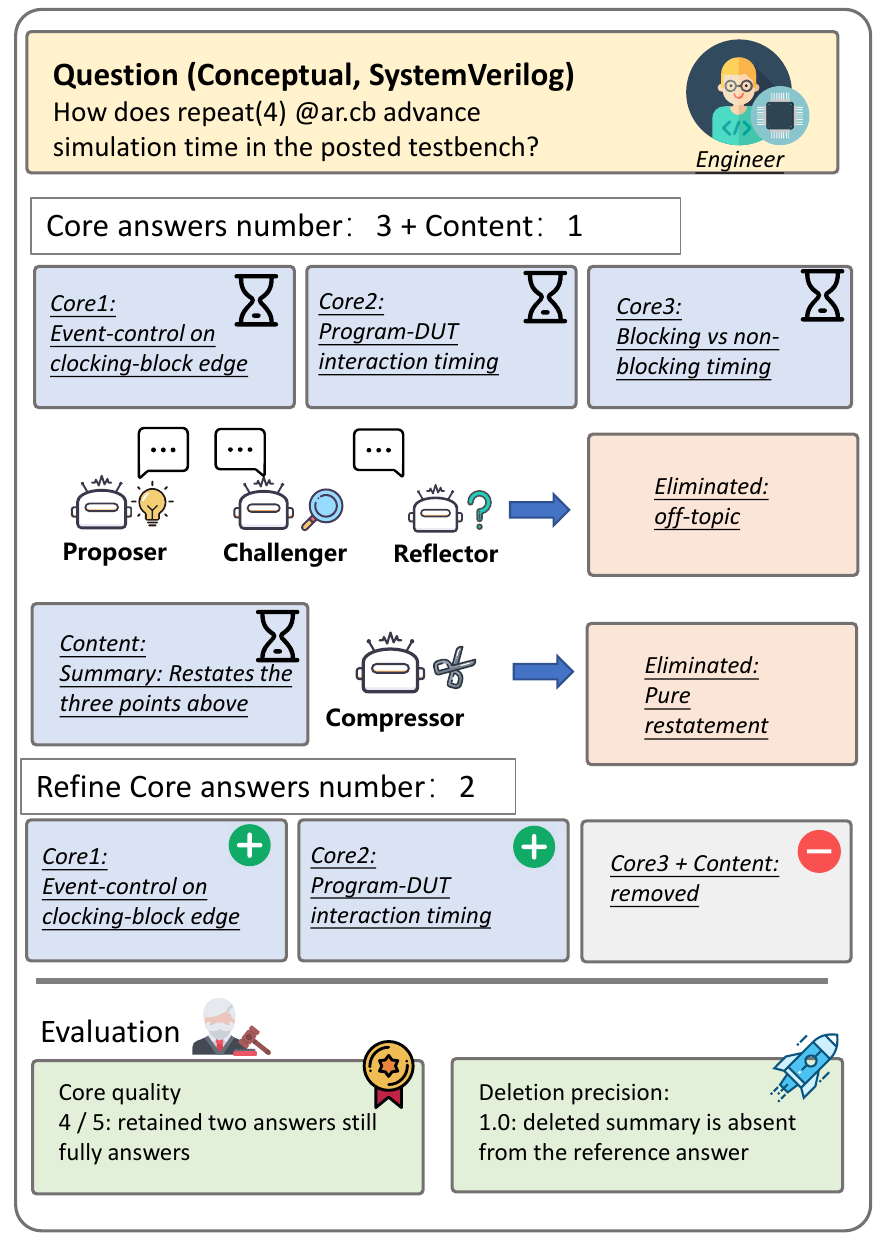}
\caption{Case study on a conceptual SystemVerilog question: the debate prunes an off-topic candidate while the compressor deletes a pure-restatement summary, raising core quality without losing substantive content.}
\label{fig:case2}
\end{figure}

\paragraph{Case 2: Removing off-topic and summary content (Conceptual).}
For a SystemVerilog question about how \texttt{repeat(4) @ar.cb} controls simulation time, the draft contained three core candidates plus a closing summary. The debate ($M_{2.2}$) kept the two on-topic explanations (event-control and program--DUT interaction) but eliminated a third candidate that digressed into blocking vs.\ non-blocking assignment timing---content the user never asked about and that does not appear in their posted code (an off-topic extra). Separately, the Compressor ($M_{2.3}$) deleted the trailing ``Summary'' block as pure restatement. Comparison against the human-accepted answer confirms both removals were appropriate: the deleted summary is absent from the reference answer (deletion precision $1.0$), and the refined core's quality score rises from $2$ to $4$ while remaining consistent with the engineer's solution. Here the verbosity reduction and the quality improvement occur together, because what was removed was off-topic or redundant rather than load-bearing.

\bibliographystyle{IEEEtran}
\bibliography{ref}